\documentclass{article}

\usepackage{spconf,amsmath,graphicx}

\usepackage{hyperref}
\usepackage{subcaption}
\usepackage{balance}
\usepackage{url}
\usepackage{booktabs}
\usepackage{breakurl}
\usepackage{amsfonts}

\title{A Large Scale Open-Source Image and Video Dataset for Robust Wildfire Detection and Classification}
\name{
\parbox{\textwidth}{
\centering
Emadeldeen Hamdan$^{1}$, Yingyi Luo$^{1}$, 
B. U{\u g}ur T\"oreyin$^{2}$, Erdem Koyuncu$^{1}$, Adam J. Watts$^{3}$\\
U{\u g}ur G\"ud\"ukbay$^{4}$, Ahmet Enis Cetin$^{1}$
}
}
\address{
    $^{1}$Department of Electrical and Computer Engineering, University of Illinois Chicago, Chicago, IL, USA\\
    $^{2}$Informatics Institute, Istanbul Technical University, \.{I}stanbul, T\"{u}rkiye\\
    $^{3}$USDA Forest Service Pacific Wildland Fire Sciences Laboratory, Washington, USA\\
    $^{4}$Department of Computer Engineering, Bilkent University, Ankara, T\"{u}rkiye
}
\begin{document}
%
\maketitle
\begin{abstract}
Wildfire detection and monitoring are critical for mitigating fire spread and reducing environmental and infrastructural damage. In this work, we introduce GWFP (Global Wildfire Prevention Dataset), a large-scale, open-source dataset of wildfire images and videos designed to support early fire and smoke detection research. GWFP contains geographically diverse wildfire scenes, including flames, smoke, Waterdog/Fog environmental conditions, Near Infrared (NIR) imagery, Ember, and challenging negative samples collected from real-world scenarios worldwide. To evaluate dataset robustness and cross-domain generalization, we benchmark multiple convolutional and transformer-based architectures across both in-domain and cross-dataset settings. Additionally, we explore lightweight frequency--spatial feature interaction using Hadamard-enhanced residual connections (HTE-ResNet) to analyze representation robustness under domain-shift conditions. Experimental results demonstrate strong cross-dataset generalization and practical utility for real-world wildfire monitoring applications. Dataset is available in Kaggle at~\url{https://www.kaggle.com/datasets/emadeldeenhamdan/wildfire}.
\end{abstract}
\section{Introduction}
\label{sec:intro}

The increasing frequency and severity of wildfires have highlighted the need for advanced technologies capable of detecting fires at their earliest stages. This need is particularly critical in regions where wildfires threaten human lives, wildlife, and infrastructure. Image and video–based wildfire detection using deep neural networks (DNNs)~\cite{AslanGTCICASSP19} has emerged as a critical tool for early fire monitoring and rapid response, becoming an increasingly active area of research in computer vision~\cite{ramos2024computer}. 

Despite the success of DNN-based approaches, their effectiveness is strongly constrained by the limited availability of large-scale, domain-specific wildfire datasets. Several studies~\cite{sousa2020wildfire} highlight data scarcity as a primary obstacle to training robust and generalizable models, often leading to overfitting and poor cross-domain performance. As a workaround, transfer learning~\cite{bozinovski2020reminder} is frequently employed, where models pre-trained on large generic image datasets are fine-tuned for wildfire detection. While beneficial, this strategy only partially mitigates the problem and does not replace the need for dedicated, diverse wildfire data.

Recent deep learning approaches for wildfire analysis have explored convolutional, transformer-based, and YOLO-style object detection frameworks for real-time fire localization and monitoring~\cite{gonccalves2024yolo}. These methods primarily focus on spatial fire localization and object detection in surveillance or aerial imagery. However, their effectiveness remains strongly dependent on the availability of diverse, well-curated wildfire datasets that support robust training and cross-domain generalization. In contrast, the primary objective of this work is the introduction of a large-scale wildfire image--video dataset designed to support comprehensive benchmarking across multiple convolutional and transformer-based architectures under diverse environmental conditions.

\begin{figure}[ht]
\centerline{\includegraphics[width=6.5cm, height=6cm]{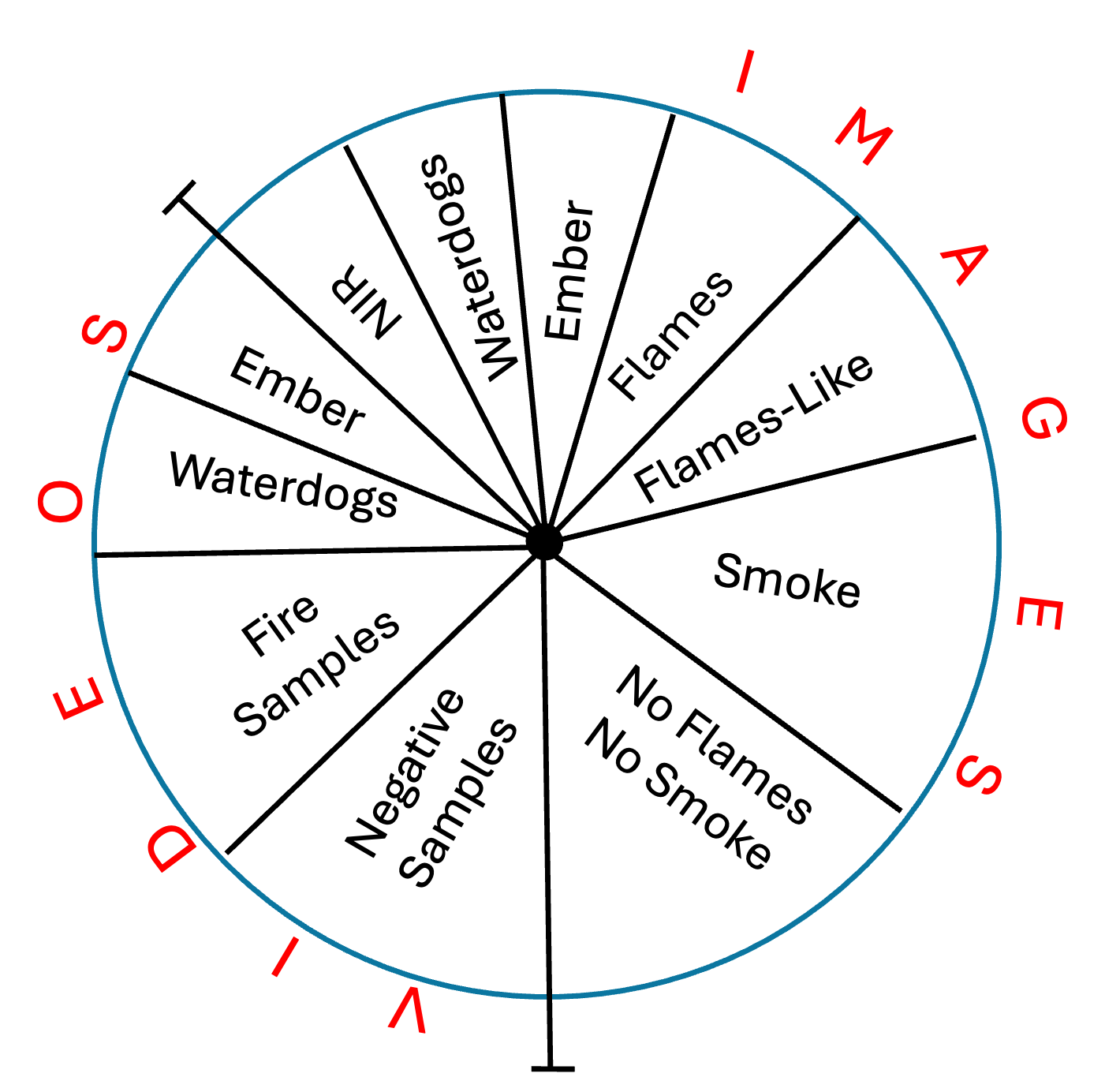}}

\caption{A detailed structure of the GWFP dataset in terms of image and video content.}
\label{fig:GWFP}
\end{figure}

To address data limitations, we introduce the Global Wildfire Prevention Dataset (GWFP), a large-scale, open-source dataset designed specifically for early wildfire detection and classification. As illustrated in Fig.~\ref{fig:GWFP}, GWFP contains both image and video samples spanning flames and smoke, embers, near-infrared (NIR) imagery, and challenging non-fire distractors including water reflections, fog, and smoke-like environmental phenomena. The data are collected from diverse geographical regions and environmental conditions to promote cross-domain generalization and robustness. GWFP is organized into well-defined classes to support scalable training and fair benchmarking across architectures.

Furthermore, to evaluate representation robustness across different architectural biases, we investigate a lightweight Hadamard-based frequency--spatial feature interaction strategy integrated into residual networks~\cite{he2016deep}, referred to as HTE-ResNet. This formulation introduces orthogonal feature mixing within skip connections with minimal computational overhead. Experimental results demonstrate improved cross-dataset robustness and highlight the practical utility of GWFP for evaluating both convolutional and transformer-based wildfire detection models. Wildfire detection is performed through patch-level classification followed by spatial localization using overlapping window-based frame analysis.

\noindent\textbf{Our main contributions are summarized as follows:}
\begin{itemize}
    \item We introduce \textbf{GWFP}, a large-scale open-source wildfire image--video dataset containing diverse real-world wildfire and environmental scenarios to support early fire and smoke detection research.

    \item We provide extensive in-domain and cross-dataset benchmarking across multiple convolutional and transformer-based architectures.

    \item We investigate lightweight Hadamard-based frequency--spatial feature interaction through HTE-ResNet to analyze representation robustness under domain-shift conditions with minimal computational overhead.
\end{itemize}

\section{Background}
\label{sec:RWork}

This section reviews publicly available wildfire and smoke detection datasets and provides a brief background on frequency–spatial feature mixing via the Hadamard transform used in our evaluation.

\subsection{The FlameVision Dataset}
\label{FlameVision}

The FlameVision dataset~\cite{jafar2023flamevision} is a publicly available wildfire and flame classification benchmark containing 8,600 RGB images collected from diverse indoor and outdoor environments, including surveillance footage and open landscapes. The dataset contains 5,000 smoke and 3,600 no-fire samples, enabling both binary and multi-class fire classification experiments.

The dataset includes both smoke and non-fire samples and is commonly used for supervised wildfire classification evaluation~\cite{elhanashi2025early}. FlameVision was selected as an external benchmark because of its different environmental conditions relative to GWFP, enabling cross-dataset evaluation of model generalization.
\subsection{MIVIA Dataset}
\label{MIVIA}
The MIVIA wildfire datasets developed by the University of Salerno include the Fire Detection Dataset, the Smoke Detection Dataset, and the Large Fire Dataset with negative samples (LFDN)~\cite{gragnaniello2025flame}. LFDN contains 36,554 images categorized into smoke, flame, combined flame--smoke, and negative environmental samples, including fire-like visual phenomena such as fog and sunlight reflections. 

However, some publicly available wildfire datasets, including MIVIA and FLAME2~\cite{9953997}, partially utilize imagery from the UCSD HPWREN camera network~\cite{hpwren2023}, which also contributes data to GWFP. To avoid potential overlap and evaluation bias, these datasets were excluded from cross-dataset benchmarking.

\begin{figure*}[ht]
    \centering
    \includegraphics[width=\linewidth, height=3cm]{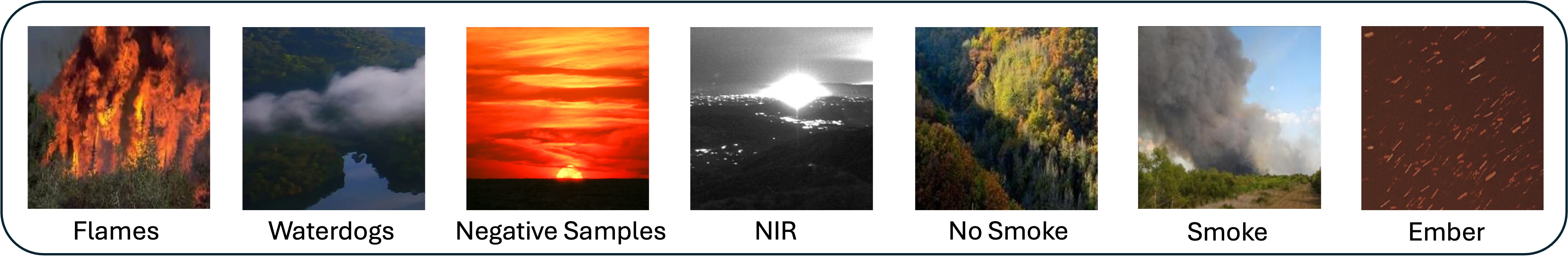}
    \caption{\textbf{GWFP Available Classes:} Representative samples from the Global Wildfire Prevention Dataset (GWFP). Classes include flame, smoke, ember, near-infrared (NIR) imagery captured using infrared cameras, and waterdogs/ fog, which are visually similar to smoke.} 
    \label{fig:gwfp_samples}
\end{figure*}
\subsection{Hadamard Transform}
\label{subsec:HT}

The Walsh--Hadamard Transform (WHT) is an orthogonal linear transform that decomposes a signal into a set of binary-valued basis functions. Owing to its computational simplicity and orthogonality, it has recently been explored in neural network architectures for efficient feature encoding and lightweight representation learning~\cite{pan2022block}.

The normalized Hadamard matrix of order $2^m$ is defined recursively as
\begin{equation}
H_m = \frac{1}{\sqrt{2}}
\begin{bmatrix}
H_{m-1} & H_{m-1} \\
H_{m-1} & -H_{m-1}
\end{bmatrix}, 
\quad H_0 = [1],
\end{equation}
where $H_m H_m^{T} = I$, indicating orthogonality. Unlike conventional convolutional filters, Hadamard-based layers operate in the transform domain and replace multiplications with sign-based additions, thereby reducing computational complexity and improving energy efficiency.

In this work, the Hadamard transform is employed as a lightweight frequency--spatial feature mixing mechanism within residual connections to evaluate representation robustness rather than as a primary architectural contribution.

\section{Methodology}
\subsection{GWFP Dataset }
\label{sec:GWFP}
With the urgent need to develop effective wildfire prevention methods, computer vision-based neural network models have emerged as a promising solution. In this work, we introduce the Global Wildfire Prevention Dataset (GWFP), a carefully curated collection of wildfire-related images and videos to support this effort. The dataset is compiled from publicly accessible sources, including HPWREN Cameras~\cite{hpwren2023} and Bilkent University. The images and videos are further sorted into classes as demonstrated in Fig.~\ref{fig:gwfp_samples}.

The Image section of our GWFP dataset (see~Table~\ref{tab:GWFP_Image}) is organized into eight classes: \textit{Flames}, \textit{Smoke}, \textit{Negative Samples}, \textit{Waterdogs/Fog}, \textit{NIR Fire}, \textit{NIR No Fire}, and \textit{Ember}. The Waterdogs are natural phenomena that resemble smoke in shape. The NIR Fire and NIR no Fire are images captured with a near-infrared (NIR) camera~\cite{qiao2024early}. 

Table~\ref{tab:GWFP_Video} presents the classes of the GWFP video section. The videos are categorized into five classes: Flame/Smoke, Negative Samples, Waterdogs, and Ember. The Flame/Smoke class includes videos featuring flames, smoke, and sequences of flames followed by smoke. Additionally, this class is separated into recordings captured by drones and stationary cameras. The Negative Samples class consists of videos with no visible flames, smoke, or fire-like occurrences, excluding water dogs. The Ember class consists of drone-captured videos of Ember caused by an active fire.

\begin{table}[ht]
    \centering
    \caption{\textbf{GWFP Image Dataset Overview:} The total image count of each class in the image section of the GWFP dataset.}
    \begin{tabular}{|l|r|}
        \hline
        \textbf{Class} & \textbf{Image Count} \\
        \hline
        Flames & 618 \\
        Smoke & 26,000 \\
        Negative Samples & 27,200 \\
        Waterdogs/Fog & 6,200 \\
        NIR Fire & 4,743 \\
        NIR No Fire & 10,060 \\
        Ember & $\approx$10,000 \\
        \hline
    \end{tabular}
    \label{tab:GWFP_Image}
\end{table}
\begin{table}[ht]
    \centering
    \caption{\textbf{GWFP Video Dataset Overview:} The total file size of each class in the video section of the GWFP dataset.}
    \begin{tabular}{|l|c|}
        \hline
        \textbf{Class} & \textbf{Count}\\
        \hline
        Flame/Smoke &161\\
        Negative Samples   &165\\
        Waterdogs/Fog  & 64 \\
        Ember   & 167\\
        \hline
    \end{tabular}
    \label{tab:GWFP_Video}
\end{table}
\begin{figure*}[t]
\centering 
\includegraphics[width=\linewidth, height=6cm]{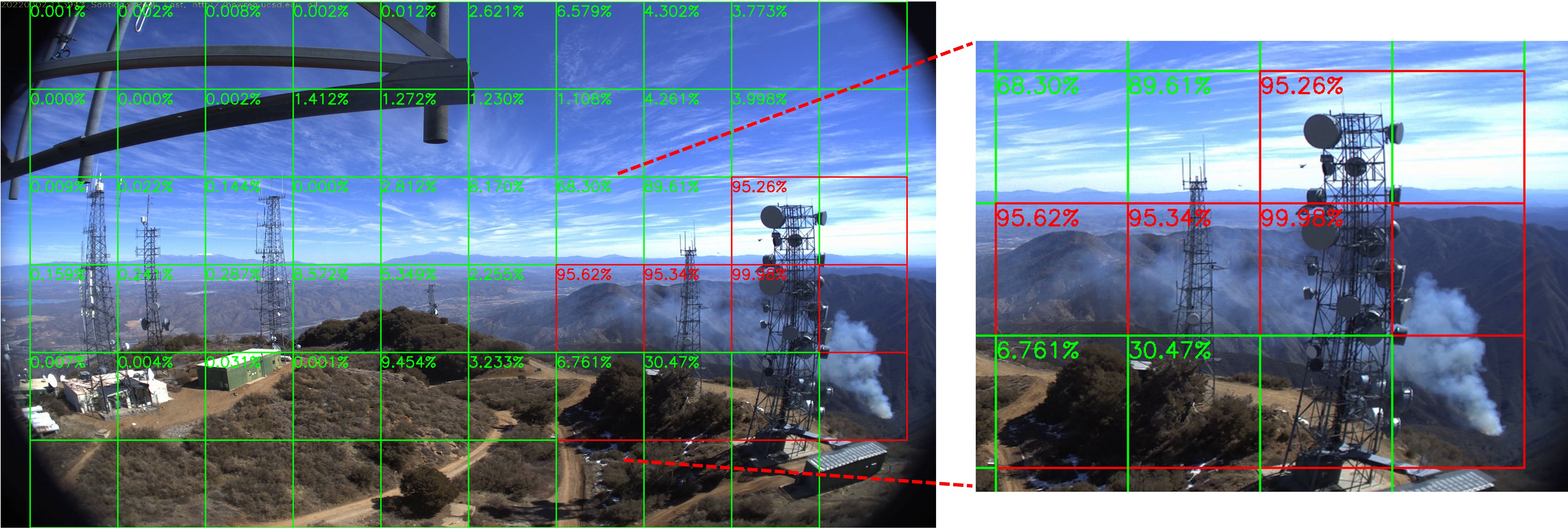} 
\caption{\textbf{Fire Detection Example:} Wildfire detection results on a GWFP video frame using the HTE-ResNet model. The frame is partitioned into overlapping $224\times224$ patches. Green boxes indicate \emph{no fire/smoke}, while red boxes indicate \emph{fire/smoke}. Numbers inside the boxes denote the predicted fire probability for each patch. Boxes without probabilities or with partial borders occur near frame edges due to the overlapping window scheme.}
\label{fig:fire_detection}
\end{figure*}

Furthermore, to facilitate robust wildfire detection, images and video frames in GWFP were processed using an overlapping sliding-window scheme that extracts $224\times224$ patches rather than resizing the entire frame. This patch-based sampling preserves local spatial detail while increasing the number of training samples. To avoid data leakage between training and validation subsets, the dataset split was performed at the image/video level prior to patch extraction, ensuring that overlapping patches originating from the same source frame were not shared across subsets. The resulting classification subset comprises three classes: 26{,}000 smoke patches, 27{,}200 negative samples, and 618 flame patches. 

The dataset was partitioned into $80\%$ for training and $20\%$ for validation. To improve robustness under class-imbalanced conditions, data augmentation techniques, including random cropping and random rotation, were applied during training, and focal loss~\cite{ross2017focal} was used to reduce the impact of class imbalance and to emphasize hard-to-classify wildfire samples.

In addition to the mentioned wildfire classification training set, we also dedicated a training dataset for Waterdogs, which can be used with more advanced models, such as motion estimation and motion-filtering-based wildfire detection~\cite{gragnaniello2025flame,Taghvaei2024}. This training set consists of 224$\times$224 processed RGB images in the first folder (Folder~A), the second frame in the second folder (Folder~B), and the label in the last folder (Folder~C). Labels are created by estimating the motion field between Folders A and B and labeling them as Waterdog or wildfire. The training set contains 5,783 images, and the testing set contains 1,447 images.

\subsection{Feature Mixing Residual Block (HTE-ResNet)}
\label{subsec:hte_resnet}

Let $\mathbf{F}_\ell \in \mathbb{R}^{C \times H \times W}$ denote the input feature tensor at layer $\ell$, 
and let $\mathcal{R}_\ell(\cdot)$ represent the residual mapping composed of convolution, normalization, 
and nonlinearity operations. The conventional residual update is defined as
\begin{equation}
\mathbf{F}_{\ell+1} = \mathbf{F}_\ell + \mathcal{R}_\ell(\mathbf{F}_\ell).
\label{eq:resnet_standard}
\end{equation}

Define the intermediate residual sum
\begin{equation}
\mathbf{S}_\ell = \mathbf{F}_\ell + \mathcal{R}_\ell(\mathbf{F}_\ell).
\label{eq:residual_sum}
\end{equation}

To incorporate orthogonal frequency--spatial mixing, we augment the update with a two-dimensional Walsh--Hadamard transform applied channel-wise to $\mathbf{S}_\ell$. The formulated feature-mixed residual update is
\begin{equation}
\mathbf{F}_{\ell+1} = \mathbf{S}_\ell + \alpha \, \mathcal{H}_{2D}(\mathbf{S}_\ell),
\label{eq:hte_residual}
\end{equation}
where $\alpha \in \mathbb{R}$ is a learnable scaling coefficient, initialized as 1, and $\mathcal{H}_{2D}(\cdot)$ denotes 
the normalized 2D Hadamard transform.

For each channel $c \in \{1,\dots,C\}$, the transform is expressed as
\begin{equation}
\mathcal{H}_{2D}(\mathbf{S}_\ell^{(c)}) 
= \mathbf{H}_H \, \mathbf{S}_\ell^{(c)} \, \mathbf{H}_W^{T},
\label{eq:ht2d}
\end{equation}
where $\mathbf{S}_\ell^{(c)} \in \mathbb{R}^{H \times W}$ and 
$\mathbf{H}_H \in \mathbb{R}^{H \times H}$, 
$\mathbf{H}_W \in \mathbb{R}^{W \times W}$ are normalized Hadamard matrices 
satisfying
\begin{equation}
\mathbf{H}_n \mathbf{H}_n^{T} = \mathbf{I}_n.
\label{eq:orthogonality}
\end{equation}

Substituting \eqref{eq:residual_sum} into \eqref{eq:hte_residual}, the final block update becomes
\begin{equation}
\mathbf{F}_{\ell+1}
= \mathbf{F}_\ell + \mathcal{R}_\ell(\mathbf{F}_\ell)
+ \alpha \, \mathcal{H}_{2D}\!\left(\mathbf{F}_\ell + \mathcal{R}_\ell(\mathbf{F}_\ell)\right).
\label{eq:final_update}
\end{equation}

Since $\mathcal{H}_{2D}(\cdot)$ consists of sign permutations and additions only, the parameter count remains unchanged, and the computational overhead is  $\mathcal{O}(CHW \log HW)$ using fast Walsh--Hadamard algorithms. Within this work, the feature-mixed residual formulation is employed as a lightweight frequency--spatial interaction mechanism to evaluate representation robustness rather than as a standalone architectural contribution.

\section{Experiments}
\label{sec:Experiments}

\subsection{Results}
\label{sec:Results}
We evaluate the proposed dataset and models on a binary fire--no-fire classification task. For this experiment, flame and smoke samples are merged into the \emph{fire} class, while flame-like distractors and negative samples are grouped into the \emph{no-fire} class. Performance is reported using Accuracy (ACC) and F1-score, with F1 used to account for class imbalance and better reflect detection reliability.

Table~\ref{tab:model_eval} summarizes cross-dataset generalization results when training on GWFP and testing on both the GWFP and FlameVision benchmarks. On the in-domain GWFP test set, most architectures achieve high performance (ACC $>92\%$ and F1 $>92\%$), indicating that the dataset supports effective supervised training. ResNet-based models show strong baseline performance, with ResNet18 and ResNet50 reaching $95.0\%$ and $94.7\%$ accuracy, respectively. The lightweight HTE-ResNet50 achieves the best in-domain performance ($\mathbf{95.2\%}$ ACC and $\mathbf{95.1\%}$ F1), demonstrating that lightweight frequency--spatial feature mixing can provide complementary gains without increasing model depth or parameter count.

Cross-dataset evaluation on FlameVision reveals a more pronounced performance spread, highlighting domain-shift effects across datasets. Transformer-based and lightweight CNN models exhibit larger drops in F1-score, suggesting reduced robustness under distribution changes. In contrast, ResNet50 maintains comparatively stable performance ($89.7\%$ ACC and $81.9\%$ F1), indicating stronger generalization among standard baselines. Notably, HTE-ResNet50 achieves the highest cross-dataset performance ($\mathbf{91.9\%}$ ACC and $\mathbf{86.4\%}$ F1), outperforming all other architectures in both metrics. This improvement suggests that the introduced orthogonal frequency--spatial mixing enhances representation diversity and mitigates overfitting to dataset-specific textures.

To further analyze temporal modeling for wildfire detection, we additionally evaluated a 3D-based spatio-temporal architecture within the primary RGB wildfire benchmark. Experimental results showed that the 3D formulation performed worse than the proposed 2D patch-based framework, likely due to limited temporal consistency across heterogeneous wildfire scenes. However, as shown in the ablation studies, temporal motion estimation was necessary to suppress false alarms in challenging Waterdog/Fog scenarios. Therefore, the motion-triggered two-stage framework was used.

Across both test sets, the proximity between ACC and F1 values indicates balanced precision--recall behavior and limited bias toward the majority class. 
Overall, the results demonstrate that GWFP supports robust model training and provides a meaningful benchmark for cross-domain evaluation. The consistent gains observed with HTE-ResNet50 further suggest that lightweight transform-domain feature interaction can improve generalization without significant computational overhead.

\begin{table}[ht]
\centering
\caption{Cross-dataset generalization performance (ACC~$|$~F1, \%) across architectures. Each cell reports Accuracy and F1-score on the corresponding test set.}
\label{tab:model_eval}
\setlength{\tabcolsep}{5pt}
\begin{tabular}{|l|c|c|}
\toprule
\textbf{Model} & \textbf{GWFP} & \textbf{FlameVision \cite{jafar2023flamevision}} \\
& ACC $|$ F1 &  ACC $|$ F1  \\
\midrule
MobileNetV2 2D~\cite{MobileNetV2}     & 92.4 $|$ 92.9 & 90.8 $|$ 85.2  \\
EfficientNetV2 2D~\cite{EfficientNetV2}  & 92.5 $|$ 92.4 & 87.4 $|$ 77.2  \\
Swin Transformer 2D~\cite{SwinTransformer} &94.6 $|$ 94.5 & 82.1 $|$ 61.2  \\
ResNet18 2D~\cite{he2016deep}        & 95.0 $|$ 94.9 & 82.3 $|$ 63.0 \\
ResNet50 2D~\cite{he2016deep}        & 94.7 $|$ 94.4 & 89.7 $|$ 81.9 \\
HTE-ResNet18 2D (Ours)& 92.2 $|$ 91.0 & 84.5 $|$ 81.5\\
\textbf{HTE-ResNet50 2D (Ours)}& \textbf{95.2 $|$ 95.1} & \textbf{91.9 $|$ 86.4}\\
\toprule
HTE-ResNet50 3D (Ours)& 93.3 $|$ 92.4 & 90.7 $|$ 87.4\\
\bottomrule
\end{tabular}
\end{table}


\subsection{Ablation Studies}
\label{subsec:ablation}

We further analyze the robustness and generalization capability of the lightweight HTE-ResNet50 evaluation across additional GWFP scenarios beyond the primary RGB wildfire detection benchmark. Table~\ref{tab:model_eval} presents preliminary evaluations on challenging subsets, including NIR wildfire detection and Waterdog/Fog environmental conditions. Although these subsets are currently smaller, the results demonstrate strong performance across different wildfire-related detection settings.

For the Waterdog/Fog subset, evaluation was performed using the motion-triggered wildfire detection framework proposed in \cite{Taghvaei2024}, which integrates neural network-based change detection into a two-stage wildfire detection pipeline. The low false alarm rate demonstrates robustness under challenging negative-only environmental conditions containing fog, atmospheric motion, and smoke-like visual patterns.

Additionally, introducing Hadamard-based feature interaction within the residual formulation improved cross-dataset stability with negligible computational overhead. The orthogonal frequency--spatial feature mixing enhances representation diversity and improves robustness under domain-shift conditions. The Ember subset is currently under expansion and will be included in future benchmark evaluations.

\begin{table}[ht]
\centering
\caption{HTE-ResNet50 performance across different GWFP evaluation tasks. ACC denotes accuracy, F1 denotes the F1-score, TDR denotes the true detection rate, and FAR denotes the false alarm rate.}
\label{tab:model_eval}
\setlength{\tabcolsep}{5pt}
\begin{tabular}{|l|c|}
\toprule
\textbf{Detection Task} & \textbf{Performance (\%)}  \\
\midrule
Waterdog/ Fog    & TDR: 77.94 $|$ FAR: 0.40  \\
NIR & ACC: 90.1 $|$ F1: 88.3  \\
Wildfire RGB & ACC: 95.2 $|$ F1: 95.1  \\

\bottomrule
\end{tabular}
\end{table}


\section{Conclusion}
\label{sec:conclusion}

This work introduced GWFP, a large-scale open-source wildfire image--video dataset designed to support early fire and smoke detection research under diverse environmental conditions. Through extensive in-domain and cross-dataset benchmarking, we demonstrated that GWFP enables robust supervised training and meaningful evaluation under domain-shift scenarios. Experimental results further indicate that dataset diversity plays a critical role in improving wildfire model generalization across heterogeneous environments. We additionally investigated lightweight Hadamard-based frequency--spatial feature interaction within residual networks as a complementary robustness evaluation strategy. The observed improvements in cross-dataset evaluation suggest that orthogonal-transform-based feature interactions can enhance representation robustness with minimal computational overhead. Future work will expand GWFP by incorporating additional wildfire-related environmental conditions and larger multimodal video subsets.

\section{Acknowledgements}
\label{sec:ack}

We thank Mr. \.{I}lhami Ayd{\i}n, General Directorate of Forestry, who provided a set of video clips. This work is funded by the Forestry Service, Pacific Northwest Research Station,  USDA, NSF POSE Grant, NSF Award 2531376 (SCC-IRG), and the original work was funded by the FIRESENSE Project, European Union, 7th Framework Program (FP-7). 

\bibliographystyle{IEEEbib}
\bibliography{strings,refs}

@article{sousa2020wildfire,
  title={Wildfire detection using transfer learning on augmented datasets},
  author={Sousa, Maria Joao and Moutinho, Alexandra and Almeida, Miguel},
  journal={Expert Systems with Applications},
  volume={142},
  note={Article no. 112975, 14~pages},
  year={2020},
  publisher={Elsevier}
}

@article{gragnaniello2025flame,
  title={{FLAME:} fire detection in videos combining a deep neural network with a model-based motion analysis},
  author={Gragnaniello, Diego and Greco, Antonio and Sansone, Carlo and Vento, Bruno},
  journal={Neural Computing and Applications},
  volume = {37}, 
  pages = {6181-6197},
  year={2025},
  publisher={Springer}
}

@misc{hpwren2023,
  author       = "{HPWREN/AI for Mankind}",
  title        = "{Wildfire Smoke Dataset}",
  year         = {2023},
  howpublished = {Available at \url{https://github.com/aiformankind/wildfire-smoke-dataset}},
  note         = {Accessed: 2026-01-15}
}

@article{qiao2024early,
  title={Early Wildfire Detection and Distance Estimation Using Aerial Visible-Infrared Images},
  author={Qiao, Linhan and Li, Shun and Zhang, Youmin and Yan, Jun},
  journal={IEEE Transactions on Industrial Electronics},
  volume={71},
  number={12},
  pages={16695-16705},
  year={2024},
  publisher={IEEE}
}

@InProceedings{AslanGTCICASSP19,
  author    = {S{\"u}leyman Aslan and
               U\u{g}ur G{\"u}d{\"u}kbay and
               B. U\u{g}ur T{\"o}reyin and
               A. Enis {\c C}etin},
  title     = {Early Wildfire Smoke Detection Based on Motion-based Geometric Image Transformation and Deep Convolutional Generative Adversarial Networks},
  booktitle = {Proceedings of IEEE International Conference on Acoustics, Speech, and Signal Processing},
  series    = {ICASSP '19}, 
  address =   {Brighton, UK},
  publisher = {IEEE},
  month =     {May},
  year      = {2019},
  pages		= {8315-8319},
  bib2html_dl_pdf = "http://www.cs.bilkent.edu.tr/~gudukbay/publications/papers/conf_papers/icassp19.pdf",
  bib2html_pubtype = {Refereed Conference Papers},
  bib2html_rescat = {Computer Vision}
}

@article{Taghvaei2024,
author = "Fatemeh Taghvaei",
title = "{Computer Vision-Based Wildfire Detection in Video: Deep Learning Using Motion Estimation}",
year = "2024",
month = "8",
journal = {University of Illinois Chciago},
url = "https://indigo.uic.edu/articles/thesis/Computer_Vision-Based_Wildfire_Detection_in_Video_Deep_Learning_Using_Motion_Estimation/27153300",
doi = "10.25417/uic.27153300.v1"}

@article{bozinovski2020reminder,
  title={Reminder of the first paper on transfer learning in neural networks, 1976},
  author={Bozinovski, Stevo},
  journal={Informatica},
  volume={44},
  number={3},
  pages = {291–302},
  year={2020}
}

@article{ramos2024computer,
  title={Computer vision for wildfire detection: a critical brief review},
  author={Ramos, Leo and Casas, Edmundo and Bendek, Eduardo and Romero, Cristian and Rivas-Echeverr{\'\i}a, Francklin},
  journal={Multimedia Tools and Applications},
  pages={1--44},
  year={2024},
  publisher={Springer}
}

@inproceedings{he2016deep,
  title={Deep residual learning for image recognition},
  author={He, Kaiming and Zhang, Xiangyu and Ren, Shaoqing and Sun, Jian},
  booktitle={Proceedings of the IEEE Conference on Computer Vision and Pattern Recognition},
  series= {CVPR~'16},
  pages={770--778},
  year={2016}
}

@article{pan2022block,
  title={{Block Walsh--Hadamard} transform-based binary layers in deep neural networks},
  author={Pan, Hongyi and Badawi, Diaa and Cetin, Ahmet Enis},
  journal={ACM Transactions on Embedded Computing Systems},
  volume={21},
  number={6},
  pages={1--25},
  year={2022},
  publisher={ACM New York, NY}
}

@article{jafar2023flamevision,
  title={FlameVision: A new dataset for wildfire classification and detection using aerial imagery},
  author={Jafar, Anam Ibn and Masud, Fatiha Binta and Ullah, Jeath Rahmat and Ahmed, Md Rayhan and others},
  journal={Mendeley Data},
  volume={4},
  year={2023}
}

@article{elhanashi2025early,
  title={Early fire and smoke detection using deep learning: A comprehensive review of models, datasets, and challenges},
  author={Elhanashi, Abdussalam and Essahraui, Siham and Dini, Pierpaolo and Saponara, Sergio},
  journal={Applied Sciences},
  volume={15},
  number={18},
  pages={10255},
  year={2025},
  publisher={MDPI}
}

@ARTICLE{9953997,
  author={Chen, Xiwen and Hopkins, Bryce and Wang, Hao and O’Neill, Leo and Afghah, Fatemeh and Razi, Abolfazl and Fulé, Peter and Coen, Janice and Rowell, Eric and Watts, Adam},
  journal={IEEE Access}, 
  title={Wildland Fire Detection and Monitoring Using a Drone-Collected RGB/IR Image Dataset}, 
  year={2022},
  volume={10},
  number={},
  pages={121301-121317},
  doi={10.1109/ACCESS.2022.3222805}
  }

@INPROCEEDINGS{MobileNetV2,
  author={Sandler, Mark and Howard, Andrew and Zhu, Menglong and Zhmoginov, Andrey and Chen, Liang-Chieh},
  booktitle={Proceedings of the  IEEE/CVF Conference on Computer Vision and Pattern Recognition}, 
  series= {CVPR~'18},
  title={{MobileNetV2:} Inverted Residuals and Linear Bottlenecks}, 
  year={2018},
  pages={4510-4520},
  doi={10.1109/CVPR.2018.00474}
  }

@InProceedings{EfficientNetV2,
  title = 	 {{EfficientNetV2:} Smaller Models and Faster Training},
  author =     {Tan, Mingxing and Le, Quoc},
  booktitle =  {Proceedings of the 38th International Conference on Machine Learning},
  pages = 	 {10096--10106},
  year = 	 {2021},
  editor = 	 {Meila, Marina and Zhang, Tong},
  volume = 	 {139},
  series = 	 {Proceedings of Machine Learning Research},
  month = 	 {18--24 Jul},
  publisher =    {PMLR},
  url = 	 {https://proceedings.mlr.press/v139/tan21a.html}
}

@INPROCEEDINGS {SwinTransformer,
author = { Liu, Ze and Lin, Yutong and Cao, Yue and Hu, Han and Wei, Yixuan and Zhang, Zheng and Lin, Stephen and Guo, Baining },
booktitle = {Proceedings of the IEEE/CVF International Conference on Computer Vision},
series= {ICCV~'21},
title = {Swin Transformer: Hierarchical Vision Transformer using Shifted Windows},
year = {2021},
pages = {9992-10002},
doi = {10.1109/ICCV48922.2021.00986},
url = {https://doi.ieeecomputersociety.org/10.1109/ICCV48922.2021.00986},
publisher = {IEEE Computer Society},
address = {Los Alamitos, CA, USA},
month =Oct}

@inproceedings{ross2017focal,
  title={Focal loss for dense object detection},
  author={Ross, T-YLPG and Doll{\'a}r, GKHP and others},
  booktitle={proceedings of the IEEE conference on computer vision and pattern recognition},
  volume={488},
  pages={2980--2988},
  year={2017}
}

@article{gonccalves2024yolo,
  title={Yolo-based models for smoke and wildfire detection in ground and aerial images},
  author={Gon{\c{c}}alves, Leon Augusto Okida and Ghali, Rafik and Akhloufi, Moulay A},
  journal={Fire},
  volume={7},
  number={4},
  pages={140},
  year={2024},
  publisher={MDPI}
}

\end{document}